\documentclass[11pt,a4paper]{article}
\usepackage[hyperref]{acl2021}
\usepackage{times}
\usepackage{latexsym}

\usepackage{microtype}

\aclfinalcopy % Uncomment this line for the final submission
\def\aclpaperid{1113} %  Enter the acl Paper ID here

\usepackage{amsfonts,amssymb}
\usepackage{threeparttable}
\usepackage{booktabs}
\usepackage{mathrsfs}
\usepackage{bm}
\usepackage{multirow}
\usepackage{float}
\usepackage{makecell}
\usepackage{mathtools}

\title{PRGC: Potential Relation and Global Correspondence Based Joint Relational Triple Extraction}
\author{\normalsize{Hengyi Zheng$^{1, 2, 3}$, Rui Wen$^3$, Xi Chen$^{3, 4}$\footnotemark[1], Yifan Yang$^3$, Yunyan Zhang$^3$} \\
\textbf{\normalsize Ziheng Zhang$^3$, Ningyu Zhang$^5$, Bin Qin$^2$\footnotemark[1], Ming Xu$^2$\footnotemark[1], Yefeng Zheng$^3$} \\
\normalsize{$^1$College of Electronics and Information Engineering, Shenzhen University} \\
\normalsize{$^2$Information Technology Center, Shenzhen University} \\
\normalsize{$^3$Tencent Jarvis Lab, Shenzhen, China} \\
\normalsize{$^4$Platform and Content Group, Tencent $^5$Zhejiang University} \\
\texttt{\small zhenghengyi2019@email.szu.edu.cn, \{qinbin,xuming\}@szu.edu.cn}\\
\texttt{\small{\{ruiwen,jasonxchen,tobyfyang,yunyanzhang,zihengzhang,yefengzheng\}@tencent.com}}
}

\date{}

\begin{document}
\maketitle
\renewcommand{\thefootnote}{\fnsymbol{footnote}}
\footnotetext[1]{Corresponding author.}
\renewcommand{\thefootnote}{\arabic{footnote}}

\begin{abstract}
% Extracting triples of relation and entities from unstructured text is a crucial task in information extraction.
Joint extraction of entities and relations from unstructured texts is a crucial task in information extraction. Recent methods achieve considerable performance but still suffer from some inherent limitations, such as redundancy of relation prediction, poor generalization of span-based extraction and inefficiency.
% {\color{blue} Some recent works achieve good performance in evaluation metrics. However, there are still some flaws in their approaches, such as relation redundancy, poor generalization of span-based extraction, inefficiency and high complexity.}
% In this paper, we propose a novel joint relational extraction framework(PRST) and decomposed the whole task into \textit{Relation Judgement}, \textit{Entity Extraction} and \textit{Subject-object Alignment} subtasks from a bran-new perspective.
In this paper, we decompose this task into three subtasks, \textit{Relation Judgement}, \textit{Entity Extraction} and \textit{Subject-object Alignment} from a novel perspective and then propose a joint relational triple extraction framework based on \textbf{P}otential \textbf{R}elation and \textbf{G}lobal \textbf{C}orrespondence (PRGC).
% PRST formulates relational triple extraction as a potential relation sequence tagging problem.
Specifically, we design a component to predict potential relations, which constrains the following entity extraction to the predicted relation subset rather than all relations;
%and apply the entity extraction task to the potential subset rather than each relation;
%then we replace the span-based extraction with sequence tagging scheme, which cannot be used in recent representative works but owns better robustness;
then a relation-specific sequence tagging component is applied to handle the overlapping problem between subjects and objects;
finally, a global correspondence component is designed to align the subject and object into a triple with low-complexity.
% considers the interaction of each token in an entity;
%{\color{blue}we replace the span-based extraction with sequence tagging scheme, which considers the interaction of each token in an entity, improves model generalization consequently}
% finally, we substitute a global correspondence scheme for previous inefficient methods to align the subject and object to the triple which is independent of relation.
% finally, instead of previous inefficient methods, we propose a global correspondence scheme which is independent of relation to 
% For one sentence, PRST proposes to predict a subset of potential relations and the global correspondence between subjects and objects for subject-object alignment, then performs sequence tagging for the subset of relations to extract the relation-specific subjects and objects. Finally, the effective pairs of subject and object are determined by the global correspondence.
Extensive experiments show that PRGC achieves state-of-the-art performance on public benchmarks with higher efficiency and delivers consistent performance gain on complex scenarios of overlapping triples.\footnote{The source code and data are released at https://github.com/hy-struggle/PRGC.}
%\revise{consistently better performance on complex scenarios of overlapping triples}.
\end{abstract}

\section{Introduction}

\begin{table}[ht]
\renewcommand\arraystretch{1.2}
\centering
\scalebox{0.68}{
\begin{tabular}{lcc}
\Xhline{1.0pt}
Subtask & Model & Component \\
\Xhline{0.6pt}
\multirow{3}{*}{Relation Judgement} & CasRel & None (Take all relations) \\
& TPLinker & None (Take all relations)  \\
& PRGC & Potential Relation Prediction\\
\Xhline{0.6pt}
\multirow{3}{*}{Entity Extraction} & CasRel & Span-based \\
& TPLinker & Span-based \\
& PRGC & Rel-Spec Sequence Tagging \\
\Xhline{0.6pt}
\multirow{3}{*}{Subject-object Alignment} & CasRel & Cascade Scheme \\
& TPLinker & Token-pair Matrix \\
& PRGC & Global Correspondence \\
%\Xhline{0.3pt}
\Xhline{1.0pt}
\end{tabular}}
\caption{Comparison of the proposed PRGC and previous methods in the respect of our new perspective with three subtasks.}
\label{improvements}
\end{table}

Identifying entity mentions and their relations which are in the form of a triple (subject, relation, object) from unstructured texts is an important task in information extraction.
Some previous works proposed to address the task with pipelined approaches which include two steps: named entity recognition~\citep{tjong-kim-sang-de-meulder-2003-introduction, ratinov-roth-2009-design} and relation prediction~\citep{10.3115/1118693.1118703, bunescu:emnlp05, DBLP:journals/corr/abs-1712-05191, DBLP:journals/corr/abs-2009-09841}.
% Generally, this can be separated as two subtasks, named entity recognition~\citep{tjong-kim-sang-de-meulder-2003-introduction, ratinov-roth-2009-design} and relation prediction~\citep{10.3115/1118693.1118703, bunescu:emnlp05}.
Recent end-to-end methods, which are based on either multi-task learning \cite{DBLP:journals/corr/abs-1909-03227} or single-stage framework~\cite{wang2020tplinker}, achieved promising performance and proved their effectiveness, but lacked in-depth study of the task.
%Extracting relational triples which are in the form of (subject, relation, object) from natural language text is a key part in automatic knowledge base construction\cite{Takanobu_Zhang_Liu_Huang_2019}.

To better comprehend the task and advance the state of the art, we propose a novel perspective to decompose the task into three subtasks: \textit{i) Relation Judgement} which aims to identify relations in a sentence, \textit{ii) Entity Extraction} which aims to extract all subjects and objects in the sentence and \textit{iii) Subject-object Alignment} which aims to align the subject-object pair into a triple.
% in public datasets, prove the effectiveness of end-to-end relational triple extraction approach.
%However, these methods also suffer from some limitations.
%As an important task in information extraction, joint relational triple extraction aims %at identifying entity mentions and their relations from unstructured texts in the triple %format (i.e., $<subject, relation, object>$).
%The existing approaches to address joint relational triple extraction can be divided %into two categories: pipelined approaches and end-to-end approaches.
%The pipelined approaches consider the task as first extracting the entities and then %identifying the relation between entities~\citep{10.3115/1118693.1118703, %chan2011exploiting}.
%The end-to-end approaches, on the other hand, are usually based on either multi-task %learning~\citep{Bekoulis_2018, DBLP:journals/corr/abs-1909-03227} or single-stage %framework~\citep{zheng-etal-2017-joint, wang2020tplinker}, which have achieved promising %performance but still suffer from some limitations,
%such as redundancy caused by relation judgement, poor generalization of span-based %entity extraction and inefficiency of subject-object alignment.
% We therefore propose to consider the task from three novel and fine-grained perspectives, \textit{Relation Judgment}, \textit{Entity Extraction} and \textit{Subject-Object Alignment}.
On the basis, we review two end-to-end methods in Table~\ref{improvements}.
For the multi-task method named CasRel \cite{DBLP:journals/corr/abs-1909-03227}, the relational triple extraction is performed in two stages which applies object extraction to all relations.
% first identifies all possible subjects in a sentence and then predicts corresponding objects for each subject on all relations.
Obviously, the way to identify relations is redundant which contains numerous invalid operations, and the span-based extraction scheme which just pays attention to start/end position of an entity leads to poor generalization. Meanwhile, it is restricted to process one subject at a time due to its subject-object alignment mechanism, which is inefficient and difficult to deploy.
%Meanwhile, the connection between subject extraction and object extraction is ignored and due to its corresponding mechanism of subject and object, it is restricted to process one subject of one sentence at a time, which means inefficient and difficult to deploy.
% Obviously, there is numerous redundancy for a specific subject to identify corresponding objects on all relations, and the span-based extraction scheme which just pays attention to start/end position of entity leads to poor generalization. Meanwhile, it is restricted to process one subject at a time due to its two stage mechanism, which means inefficient and difficult to deploy.
For the single-stage framework named TPLinker \cite{wang2020tplinker}, in order to avoid the exposure bias in subject-object alignment, it exploits a rather complicated decoder which leads to sparse label and low convergence rate while the problems of relation redundancy and poor generalization of span-based extraction are still unsolved.

To address aforementioned issues, we propose an end-to-end framework which consists of three components: \textit{Potential Relation Prediction}, \textit{Relation-Specific Sequence Tagging} and \textit{Global Correspondence}, which fulfill the three subtasks accordingly as shown in Table~\ref{improvements}.
% we present a novel framework by a brand-new view which generally adapts to the RE task.
%In our opinion, the RE task can be decomposed into three subtasks: \textit{Relation Judgement}, \textit{Entity Extraction} and \textit{Subject-object Alignment}, as shown in Table \ref{improvements}, we apply this view to previous works to show our improvements on each of the subtask.

% PRST formulates the RE task as a potential relation sequence tagging problem with a fine-grained view and make improvements on each of the subtasks. We apply this view to previous works and compare them with our own, which is shown in Table \ref{improvements}. 
%instead of applying the entity extraction subtask to all relations, 
%we design a component which greatly reduces redundant relations by identifying potential relations.

For \textit{Relation Judgement}, we predict potential relations by the \textit{Potential Relation Prediction} component rather than preserve all redundant relations, 
which reduces computational complexity and achieves better performance, especially when there are many relations in the dataset.\footnote{For example, the WebNLG dataset \cite{gardent-etal-2017-creating} has hundreds of relations but only seven valid relations for one sentence mostly.} 
For \textit{Entity Extraction}, we use a more robust \textit{Relation-Specific Sequence Tagging} component (Rel-Spec Sequence Tagging for short) to extract subjects and objects separately, to naturally handle overlapping between subjects and objects.
% we consider the interaction of tokens in one entity, thus improve model performance;
%For \textit{Subject-object Alignment}, instead of previous inefficient methods, we design a global matrix to determine whether a specific subject-object pair is valid in a triple according to the value of the corresponding position in the matrix. We disassociate this component from the relation of a subject-object pair, thus increase efficiency and reduce complexity of our model.
For \textit{Subject-object Alignment}, unlike TPLinker which uses a relation-based token-pair matrix, we design a relation-independent \textit{Global Correspondence} matrix to determine whether a specific subject-object pair is valid in a triple.
% we disassociate it from relation, thus reduce complexity of model.
%This component reduces computation of model and brings higher performance, especially when there are many relations in the dataset such as the WebNLG\cite{gardent-etal-2017-creating} which has hundreds of relations but only 7 valid relations for one sentence mostly;
%On the one hand, this component reduces computation of model, especially when there are many relations in the dataset such as the WebNLG\cite{gardent-etal-2017-creating}, which has hundreds of relations but only 7 valid relations for one sentence mostly; 
%on the other hand, experiments show that this also promotes the model to be more focused on the entity extraction of potential relations rather than all relations and achieve higher performance.
% the greatly reduced redundancy of negative relations
% we greatly reduce redundant relations by predicting potential relations, thus improve the performance and efficiency simultaneously; 
%For \textit{Entity Extraction}, we use sequence tagging instead of span-based extraction to extract entities, which considers the interaction between tokens in one entity but cannot be implemented to recent representational works because of their schemes, experiments show that this scheme is more robust.

Given a sentence, PRGC first predicts a subset of potential relations and a global matrix which contains the correspondence score between all subjects and objects; then performs sequence tagging to extract subjects and objects for each potential relation in parallel; finally enumerates all predicted entity pairs, which are then pruned by the global correspondence matrix.
It is worth to note that the experiment (described in Section \ref{ana on exposure}) shows that the \textit{Potential Relation Prediction} component of PRGC is overall beneficial, even though it introduces the exposure bias that is usually mentioned in prior single-stage methods to prove their advantages.
%Experimental results show that PRST not only solve the problem of overlapping entities and relations(shown in Figure \ref{overlapping}) which is usually mentioned in prior works\cite{nayak2019effective}\cite{DBLP:journals/corr/abs-1909-03227}\cite{yuan2020relation}\cite{wang2020tplinker}, but also outperforms previous works and achieves the state-of-the-art results on several benchmarks with higher efficiency and fewer decoder parameters.
%Experimental results show that PRST not only solve the problem of complex scenes such as overlapping patterns which is usually mentioned in prior works\cite{nayak2019effective}\cite{DBLP:journals/corr/abs-1909-03227}\cite{yuan2020relation}\cite{wang2020tplinker}, but also outperforms the state-of-the-art works on public benchmarks with higher efficiency and fewer decoder parameters.
%Experimental results show that PRGC not only solves the problem of complex scenes such as overlapping patterns which contains the \textit{SEO(Single Entity Overlap)}, \textit{EPO(Entity Pair Overlap)} and \textit{SOO(Subject Object Overlap)} types\footnote{More details about overlapping patterns are shown in Appendix \ref{overlapping appendix}.}, but also outperforms the state-of-the-art methods on public benchmarks with higher efficiency and less parameters. 

Experimental results show that PRGC outperforms the state-of-the-art methods on public benchmarks with higher efficiency and fewer parameters. Detailed experiments on complex scenarios such as various overlapping patterns, which contain the \textit{Single Entity Overlap (SEO)}, \textit{Entity Pair Overlap (EPO)} and \textit{Subject Object Overlap (SOO)} types\footnote{More details about overlapping patterns are shown in Appendix \ref{overlapping appendix}.} show that our method owns consistent advantages.
The main contributions of this paper are as follows:
%It is worth to note that one of the most important advantage of single-stage model compared with joint learning model is avoiding the influence of exposure bias\cite{wang2020tplinker}, but our experimental results(described in Section \ref{ana on exposure}) show that the relation judgement component of our model which introduces the exposure bias is more beneficial than harmful.
%which is usually mentioned in prior works\cite{nayak2019effective}\cite{DBLP:journals/corr/abs-1909-03227}\cite{yuan2020relation}\cite{wang2020tplinker}
%\newpage
\begin{enumerate}
    %\item We tackle the relational triple extraction task from a novel perspective which is generally suitable for previous methods as shown in Table \ref{improvements} and decomposes the task into three subtasks: \textit{Relation Judgement}, \textit{Entity Extraction} and \textit{Subject-object Alignment}.
    \item We tackle the relational triple extraction task from a novel perspective which decomposes the task into three subtasks: \textit{Relation Judgement}, \textit{Entity Extraction} and \textit{Subject-object Alignment}, and previous works are compared on the basis of the proposed paradigm as shown in Table \ref{improvements}.
    
    \item Following our perspective, we propose a novel end-to-end framework and design three components with respect to the subtasks which greatly alleviate the problems of redundant relation judgement, poor generalization of span-based extraction and inefficient subject-object alignment, respectively.
    
    \item We conduct extensive experiments on several public benchmarks, which indicate that our method achieves state-of-the-art performance, especially for complex scenarios of overlapping triples. Further ablation studies and analyses confirm the effectiveness of each component in our model.
    
    \item In addition to higher accuracy, experiments show that our method owns significant advantages in complexity, number of parameters, floating point operations (FLOPs) and inference time compared with previous works.

\end{enumerate}

\section{Related Work}
Traditionally, relational triple extraction has been studied as two separated tasks: entity extraction and relation prediction. Early works~\citep{10.3115/1118693.1118703, chan2011exploiting} apply the pipelined methods to perform relation classification between entity pairs after extracting all the entities. 
%To ease the problems of error propagation and neglect between the two stage in pipelined methods, end-to-end model has attracted much attention.
To establish the correlation between these two tasks, joint models have attracted much attention. Prior feature-based joint models~\citep{yu-lam-2010-jointly, li-ji-2014-incremental, miwa-sasaki-2014-modeling, ren2017cotype} require a complicated process of feature engineering and rely on various NLP tools with cumbersome manual operations.

\begin{figure*}[ht]
\centering
\includegraphics[width=\textwidth]{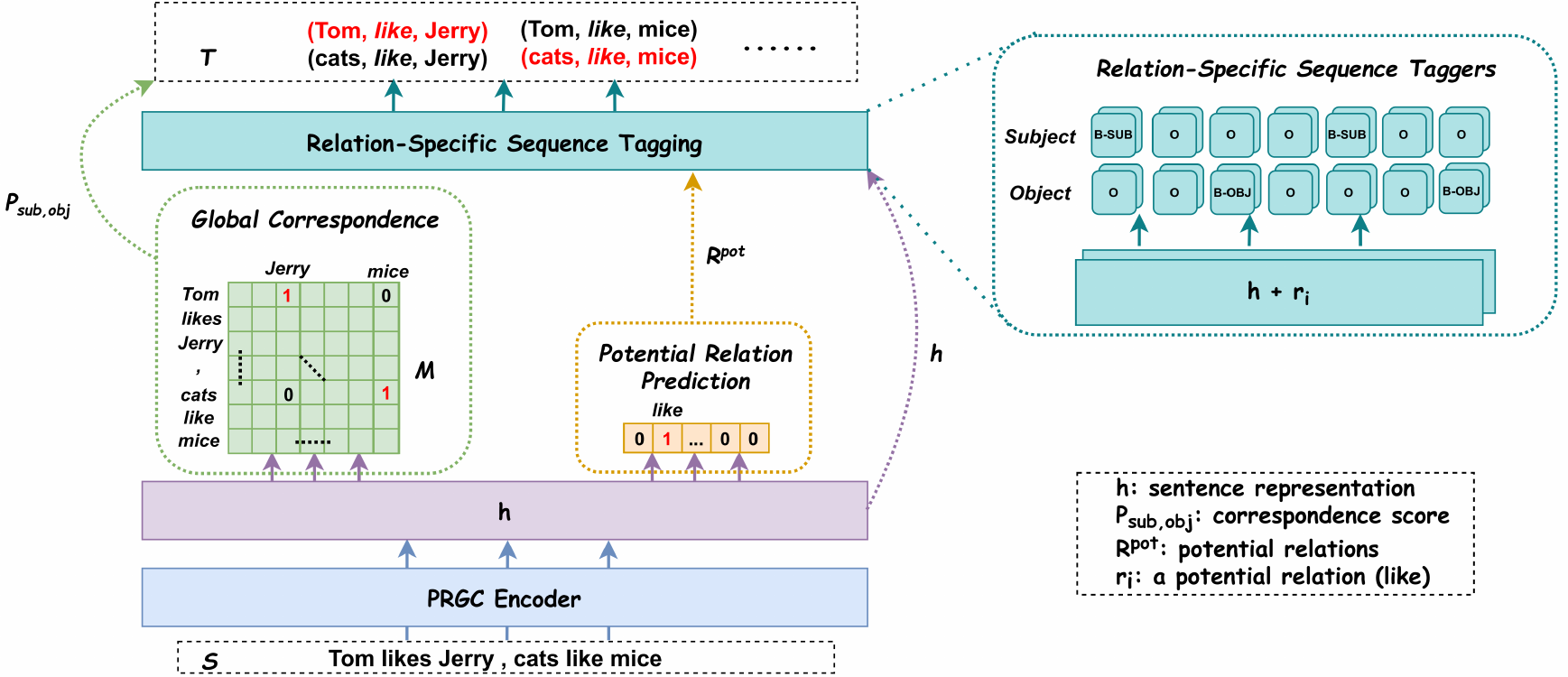}
\caption{The overall structure of PRGC. Given a sentence $S$, PRGC predicts a subset of potential relations $R^{pot}$ and a global correspondence $M$ which indicates the alignment between subjects and objects.
Then for each potential relation, a relation-specific sentence representation is constructed for sequence tagging. 
Finally we enumerate all possible subject-object pairs and get four candidate triples for this particular example, but only two triples are left (marked red) after applying the constraint of global correspondence.}
\label{overview}
\end{figure*}

Recently, the neural network model which reduces manual involvement occupies the main part of the research. \citet{zheng-etal-2017-joint} proposed a novel tagging scheme that unified the role of the entity and the relation between entities in the annotations, thus the joint extraction task was converted to a sequence labeling task but it failed to solve the overlapping problems. \citet{Bekoulis_2018} proposed to first extract all candidate entities, then predict the relation of every entity pair as a multi-head selection problem, which shared parameters but did not decode jointly. \citet{nayak2019effective} employed an encoder-decoder architecture and a pointer network based decoding approach where an entire triple was generated at each time step.

To handle the problems mentioned above,
\citet{DBLP:journals/corr/abs-1909-03227} presented a cascade framework, which first identified all possible subjects in a sentence, then for each subject, applied span-based taggers to identify the corresponding objects based on each relation. This method leads to redundancy on relation judgement, and is not robust due to the span-based scheme on entity extraction. Meanwhile, the alignment scheme of subjects and objects limits its parallelization. 
In order to represent the relation of triple explicitly, \citet{yuan2020relation} presented a relation-specific attention to assign different weights to the words in context under each relation, but it applied a naive heuristic nearest neighbor principle to combine the entity pairs which means the nearest subject and object entities will be combined into a triple. This is obviously not in accordance with intuition and fact. Meanwhile, it is also redundant on relation judgement. 
The state-of-the-art method named TPLinker \cite{wang2020tplinker} employs a token pair linking scheme which performs two $O(n^2)$ matrix operations for extracting entities and aligning subjects with objects under each relation of a sentence, causing extreme redundancy on relation judgement and complexity on subject-object alignment, respectively. And it also suffers from the disadvantage of span-based extraction scheme.

\section{Method}

In this section, we first introduce our perspective of relational triple extraction task with a principled problem definition, then elaborate each component of the PRGC model. An overview illustration of PRGC is shown in Figure \ref{overview}.
% description of BERT\cite{devlin2019bert} encoder is shown in Appendix \ref{bert}. 

\subsection{Problem Definition}
\label{problem definition}
The input is a sentence $S = \{x_1, x_2, ..., x_n\}$ with $n$ tokens. The desired outputs are relational triples as $T(S) = \{(s, r, o) | s, o \in E, r \in R \}$, where $E$ and $R$ are the entity and relation sets, respectively. In this paper, the problem is decomposed into three subtasks:

\paragraph{Relation Judgement} For the given sentence $S$, this subtask predicts potential relations it contains. The output of this task is $Y_r(S) = \{r_1, r_2, ..., r_m | r_i \in R\}$, where $m$ is the size of potential relation subset.

\paragraph{Entity Extraction} For the given sentence $S$ and a predicted potential relation $r_i$, this subtask identifies the tag of each token with BIO (i.e., Begin, Inside and Outside) tag scheme~\citep{tjong-kim-sang-veenstra-1999-representing, ratinov-roth-2009-design}.
Let $t_j$ denote the tag. The output of this task is $Y_e(S, r_i | r_i \in R) = \{t_1, t_2, ..., t_n \}$.

\paragraph{Subject-object Alignment} For the given sentence $S$, this subtask predicts the correspondence score between the start tokens of subjects and objects. That means only the pair of start tokens of a true triple has a high score, while the other token pairs have a low score. Let $\mathbf{M}$ denote the global correspondence matrix. The output of this task is $Y_s(S) = \mathbf{M} \in \mathbb{R}^{n \times n}$.
%the higher the score, the higher the confidence that an entity pair belongs to a triple. 

% \paragraph{Subject-object Alignment} For the given sentence $S$ and enumeration of arbitrary subject-object entity pair, to predict subject-object alignment score, the higher the score, the higher the confidence that the entity pair belongs to a triple, and vice versa. Let $\mathbf{M}$ denote the subject-object alignment score matrix, the output of this task is $Y_g(S) = \mathbf{M} \in \mathbb{R}^{n \times n}$.

\subsection{PRGC Encoder}
The output of PRGC Encoder is $Y_{enc}(S) = \{h_1, h_2, ..., h_n | h_i \in \mathbb{R}^{d \times 1}\}$, where $d$ is the embedding dimension, and $n$ is the number of tokens. 
We use a pre-trained BERT model\footnote{Please refer to the original paper~\cite{devlin2019bert} for detailed descriptions.} \cite{devlin2019bert} to encode the input sentence for a fair comparison, but theoretically it can be extended to other encoders, such as Glove~\cite{pennington-etal-2014-glove} and RoBERTa~\cite{DBLP:journals/corr/abs-1907-11692}.

%We use a pre-trained BERT model \cite{devlin2019bert} to encode the input sentence. The output of BERT is $Y_{BERT}(S) = \{v_1, v_2, ..., v_n | v_i \in \mathbb{R}^{d \times 1}\}$, where $d$ is the embedding size, and $n$ is the number of tokens. 

\subsection{PRGC Decoder}
%In this section, we will describe the instantiation of PRGC decoder that follows the BERT\cite{devlin2019bert} encoder\footnote{Description of BERT encoder is shown in Appendix \ref{bert}.}, which consists of three main components, corresponding to the three subtasks described in Section \ref{problem definition}.
In this section, we describe the instantiation of PRGC decoder that consists of three components.

\subsubsection{Potential Relation Prediction}
\label{relation judgement}
This component is shown as the orange box in Figure \ref{overview} where $R^{pot}$ is the potential relations.
Different from previous works~\citep{DBLP:journals/corr/abs-1909-03227, yuan2020relation, wang2020tplinker} which redundantly perform entity extraction to every relation, given a sentence, we first predict a subset of potential relations that possibly exist in the sentence, and then the entity extraction only needs to be applied to these potential relations. 
%On the one hand, this greatly reduces computation, especially when there are many relations, such as the WebNLG\footnote{Following prior works, we use two versions of WebNLG dataset, and it has 171 and 216 relations respectively.}\cite{gardent-etal-2017-creating} dataset, which has hundreds of relations but only 7 effective relations for one sentence mostly;
%on the other hand, as the ablation study shown in Table \ref{ablation}, the greatly reduced redundancy of negative relations also promotes the model to be more focused on the entity extraction of potential relations and achieve higher performance.
Given the embedding $\mathbf{h} \in \mathbb{R}^{n \times d}$ of a sentence with $n$ tokens, each element of this component is obtained as:
% The value of this component is obtained as follows:
\begin{equation}
\small
\begin{split}
\begin{aligned}
\label{rel_eq}
    \mathbf{h}^{avg} &= Avgpool(\mathbf{h}) \in \mathbb{R}^{d \times 1} \\
    P_{rel} &= \sigma(\mathbf{W}_r\mathbf{h}^{avg} + \mathbf{b}_r)
\end{aligned}
\end{split}
\end{equation}
where $Avgpool$ is the average pooling operation \cite{lin2013network}, $\mathbf{W}_r \in \mathbb{R}^{d \times 1}$ is a trainable weight and $\sigma$ denotes the sigmoid function.

We model it as a multi-label binary classification task, and the corresponding relation will be assigned with tag 1 if the probability exceeds a certain threshold $\lambda_1$ or with tag 0 otherwise (as shown in Figure \ref{overview}), so next we just need to apply the relation-specific sequence tagging to the predicted relations rather than all relations.

\subsubsection{Relation-Specific Sequence Tagging}
\label{sequence tagging}
As shown in Figure \ref{overview}, we obtain several relation-specific sentence representations of potential relations described in Section \ref{relation judgement}. Then, we perform two sequence tagging operations to extract subjects and objects, respectively.
%The reason why we divide the sequence tagging into two parts is to handle the special overlapping pattern named \textit{SubjectObjectOverlap(SOO)}, so we can simplify it to one sequence tagging operation with two types of entities if there are no \textit{SOO} pattern in the dataset\footnote{For example, the \textit{SOO} pattern is rare in the NYT\cite{Riedel2010ModelingRA} dataset.}. Detailed operations of this component on each token are as follows:
The reason why we extract subjects and objects separately is to handle the special overlapping pattern named \textit{Subject Object Overlap (SOO)}. We can also simplify it to one sequence tagging operation with two types of entities if there are no \textit{SOO} patterns in the dataset.\footnote{For example, the \textit{SOO} pattern is rare in the NYT \cite{Riedel2010ModelingRA} dataset.}

For the sake of simplicity and fairness, we abandon the traditional LSTM-CRF \cite{DBLP:journals/corr/HuangXY15} network but adopt the simple fully connected neural network. Detailed operations of this component on each token are as follows:
%The reason why we extract subject and object separately is to handle the special overlapping pattern named \textit{SubjectObjectOverlap(SOO)}, so we can simplify it to one sequence tagging operation with two types of entities if there are no \textit{SOO} pattern in the dataset\footnote{For example, the \textit{SOO} pattern is rare in the NYT\cite{Riedel2010ModelingRA} dataset.}. For the sake of simplicity and efficiency, we abandon the traditional lstm-crf network for sequence tagging, but adopt the extremely simple linear neural network, detailed operations of this component on each token are as follows:
\begin{equation}
\small
\begin{split}
\label{seq_eq}
    \mathbf{P}_{i, j}^{sub} = Softmax(\mathbf{W}_{sub}(\mathbf{h}_i + \mathbf{u}_j) + \mathbf{b}_{sub}) \\
    \mathbf{P}_{i, j}^{obj} = Softmax(\mathbf{W}_{obj}(\mathbf{h}_i + \mathbf{u}_j) + \mathbf{b}_{obj})
\end{split}
\end{equation}
where $\mathbf{u}_j \in \mathbb{R}^{d \times 1}$ is the $j$-th relation representation in a trainable embedding matrix $\mathbf{U} \in \mathbb{R}^{d \times n_r}$ where $n_r$ is the size of full relation set, $\mathbf{h}_i \in \mathbb{R}^{d \times 1}$ is the encoded representation of the $i$-th token, and $\mathbf{W}_{sub}, \mathbf{W}_{obj} \in \mathbb{R}^{d \times 3}$ are trainable weights where the size of tag set \{B, I, O\} is 3.

\begin{table*}[ht]
\renewcommand\arraystretch{1.2}
\centering
\scalebox{0.8}{
\begin{tabular}{lcccccccccccc}
\Xhline{1.0pt}
\multirow{2}{*}{Dataset} & \multicolumn{3}{c}{\#Sentences} & \multicolumn{8}{c}{Details of test set} \\
\cmidrule(r){2-4}\cmidrule(r){5-12} & Train & Valid & Test & Normal & SEO & EPO & SOO & $N = 1$ & $N > 1$ & \#Triples & \#Relations   \\
\Xhline{0.6pt}
NYT* & 56,195 & 4,999 & 5,000  & 3,266 & 1,297 & 978 & 45 & 3,244 & 1,756  & 8,110& 24\\
WebNLG* & 5,019 & 500 & 703  & 245 & 457 & 26 & 84 & 266 & 437   & 1,591 & 171 \\
NYT & 56,196 & 5,000 & 5,000  & 3,071 & 1,273 & 1,168 & 117 & 3,089 & 1,911  & 8,616  &24\\
WebNLG & 5,019 & 500 & 703  & 239 & 448 & 6 & 85 & 256 & 447  & 1,607 &216\\
\Xhline{1.0pt}
\end{tabular}}
\caption{Statistics of datasets used in our experiments where $N$ is the number of triples in a sentence. Note that one sentence can have SEO, EPO and SOO overlapping patterns simultaneously, and the relation set of WebNLG is bigger than WebNLG*.}
\label{statistics}
\end{table*}

\subsubsection{Global Correspondence}
\label{global co}
After sequence tagging, we acquire all possible subjects and objects with respect to a relation of the sentence, then we use a global correspondence matrix to determine the correct pairs of the subjects and objects. 
It should be noted that the global correspondence matrix can be learned simultaneously with potential relation prediction since it is independent of relations.
The detailed process is as follows: first we enumerate all the possible subject-object pairs; then we check the corresponding score in the global matrix for each pair, retain it if the value exceeds a certain threshold $\lambda_2$ or filter it out otherwise.
%but we don't know which pairs of a subject and a object should be combined. To solve this problem, we propose a global correspondence matrix to determine whether a specific subject-object pair should be combined into a triple according to the value of the corresponding position in the matrix.
%\citet{zheng-etal-2017-joint} and \citet{yuan2020relation} apply a naive scheme named the heuristic nearest principle to combine entity pairs, which means the nearest subject and object entities will be combined into a triple, but this is obviously not in accordance with intuition and fact. In order to solve this problem, we propose a global alignment matrix to determine whether a specific subject-object pair should be combined into a triple according to the value of the corresponding position in the matrix.

As shown in the green matrix $M$ in Figure \ref{overview}, given a sentence with $n$ tokens, the shape of global correspondence matrix will be $\mathbb{R}^{n \times n}$. 
Each element of this matrix is about the start position of a paired subject and object, which represents the confidence level of a subject-object pair, the higher the value, the higher the confidence level that the pair belongs to a triple.
For example, the value about ``\textit{Tom}" and ``\textit{Jerry}" at row 1, column 3 will be high if they are in a correct triple such as ``\textit{(Tom, like, Jerry)}".
%and it will be set to 1 for a multi-label binary classification task. 
The value of each element in the matrix is obtained as follows:
% And each position of the matrix indicates the start position of a paired subject and object entity.
\begin{equation}
\label{glo_eq}
\small
P_{i_{sub}, j_{obj}} = \sigma(\mathbf{W}_g[\mathbf{h}^{sub}_i; \mathbf{h}^{obj}_j] + \mathbf{b}_g)
\end{equation}
%``$;$" is the concat operation,
where $\mathbf{h}^{sub}_i, \mathbf{h}^{obj}_j \in \mathbb{R}^{d \times 1}$ are the encoded representation of the $i$-th token and $j$-th token in the input sentence forming a potential pair of subject and object, $\mathbf{W}_g \in \mathbb{R}^{2d \times 1}$ is a trainable weight, and $\sigma$ is the sigmoid function. 
%Note that the correspondence score of all the triples in one sentence is contained in the global matrix, which is relation independent.

\subsection{Training Strategy}
\label{loss func}
We train the model jointly, optimize the combined objective function during training time and share the parameters of the PRGC encoder. The total loss can be divided into three parts as follows:
\begin{equation}
\small
\begin{aligned}
    \mathcal{L}_{rel} = -\frac{1}{n_r}\sum_{i=1}^{n_r}(y_{i}\log{P_{rel}}+(1-y_i)\log{(1-P_{rel}))}
\end{aligned}
\end{equation}
\begin{equation}
\small
\begin{aligned}
    \mathcal{L}_{seq} = -\frac{1}{2 \times n \times n_r^{pot}}\sum_{t \in \{sub, obj\}}\sum_{j=1}^{n_r^{pot}}\sum_{i=1}^{n}\mathbf{y}_{i, j}^{t}\log{\mathbf{P}_{i, j}^{t}}
\end{aligned}
\end{equation}
\begin{equation}
\small
\begin{aligned}
    \mathcal{L}_{global} = &-\frac{1}{n^2}\sum_{i=1}^{n}\sum_{j=1}^{n}(y_{i, j}\log{P_{i_{sub}, j_{obj}}}\\
    & +(1-y_{i, j})\log{(1-P_{i_{sub}, j_{obj}})})
\end{aligned}
\end{equation}
where $n_r$ is the size of full relation set and $n_r^{pot}$ is the size of potential relation subset of the sentence.
The total loss is the sum of these three parts,
\begin{equation}
\small
    \mathcal{L}_{total} = \alpha\mathcal{L}_{rel} + \beta\mathcal{L}_{seq} + \gamma\mathcal{L}_{global}.
\end{equation}
Performance might be better by carefully tuning the weight of each sub-loss, but we just assign equal weights for simplicity (i.e., $\alpha = \beta = \gamma = 1$).

\begin{table*}[ht]
\renewcommand\arraystretch{1.2}
\centering
\scalebox{0.78}{
\begin{tabular}{lcccccccccccc}
\Xhline{1.0pt}
% 49, 50, 54, 55
\multirow{2}{*}{Model} & \multicolumn{3}{c}{NYT*} & \multicolumn{3}{c}{WebNLG*} & \multicolumn{3}{c}{NYT} & \multicolumn{3}{c}{WebNLG} \\
\cmidrule(r){2-4}\cmidrule(r){5-7}\cmidrule(r){8-10}\cmidrule(r){11-13} 
& Prec. & Rec. & F1 & Prec. & Rec. & F1 & Prec. & Rec. & F1 & Prec. & Rec. & F1 \\
\Xhline{0.6pt}
NovelTagging \cite{zheng-etal-2017-joint} & - & - & - & - & - & - & 32.8 & 30.6 & 31.7 & 52.5 & 19.3 & 28.3 \\
CopyRE \cite{zeng-etal-2018-extracting} & 61.0 & 56.6 & 58.7 & 37.7 & 36.4 & 37.1 & - & - & - & - & - & -    \\
MultiHead \cite{Bekoulis_2018} & - & - & - & - & - & - & 60.7 & 58.6 & 59.6 & 57.5 & 54.1 & 55.7   \\
GraphRel \cite{fu-etal-2019-graphrel} & 63.9 & 60.0 & 61.9 & 44.7 & 41.1 & 42.9 & - & - & - & - & - & - \\
OrderCopyRE \cite{zeng-etal-2019-learning} & 77.9 & 67.2 & 72.1 & 63.3 & 59.9 & 61.6 & - & - & - & - & - & - \\
ETL-span \cite{yu2020joint} & 84.9 & 72.3 & 78.1 & 84.0 & 91.5 & 87.6 & 85.5 & 71.7 & 78.0 & 84.3 & 82.0 & 83.1 \\
WDec \cite{nayak2019effective} & \textbf{94.5} & 76.2 & 84.4 & - & - & - & - & - & - & - & - & - \\
RSAN$\ddagger$ \cite{yuan2020relation} & - & - & - & - & - & - & 85.7 & 83.6 & 84.6 & 80.5 & 83.8 & 82.1 \\
% CasRel$_{LSTM}\ddagger$\cite{DBLP:journals/corr/abs-1909-03227} & 84.2 & 83.0 & 83.6 & 86.9 & 80.6 & 83.7 & - & - & - & - & - & - \\
CasRel$_{Random}\ddagger$ \cite{DBLP:journals/corr/abs-1909-03227} & 81.5 & 75.7 & 78.5 & 84.7 & 79.5 & 82.0 & - & - & - & - & - & - \\
CasRel$_{BERT}\ddagger$ \cite{DBLP:journals/corr/abs-1909-03227} & 89.7 & 89.5 & 89.6 & \underline{93.4} & 90.1 & 91.8 & - & - & - & - & - & - \\
% TPLinker$_{LSTM}\ddagger$\cite{wang2020tplinker} & 83.8 & 83.4 & 83.6 & 90.8 & 90.3 & 90.5 & 86.0 & 82.0 & 84.0 & \textbf{91.9} & 81.6 & 86.4 \\
% TPLinker$_{Random}\S$\cite{wang2020tplinker} & 89.5 & 85.6 & 87.5 & 88.5 & 78.2 & 83.0 & 90.0 & 87.0 & 88.5 & - & - & - \\
TPLinker$_{BERT}\ddagger$ \cite{wang2020tplinker} & 91.3 & \textbf{92.5} & \underline{91.9} & 91.8 & \underline{92.0} & \underline{91.9} & \underline{91.4} & \textbf{92.6} & \underline{92.0} & \underline{88.9} & \underline{84.5} & \underline{86.7} \\
\Xhline{0.6pt}
% FAST$_{LSTM}$ & 93.3 & 91.9 & 92.6 & 92.8 & 92.1 & 92.4 & 93.5 & 91.9 & 92.7 & 88.6 & 87.1 & 87.8 \\
PRGC$_{Random}$ & 89.6 & 82.3 & 85.8 & 90.6 & 88.5 & 89.5 & 87.8 & 83.8 & 85.8 & 82.5 & 79.2 & 80.8 \\
PRGC$_{BERT}$ & \underline{93.3} & \underline{91.9} & \textbf{92.6} & \textbf{94.0} & \textbf{92.1} & \textbf{93.0} & \textbf{93.5} & \underline{91.9} & \textbf{92.7} & \textbf{89.9} & \textbf{87.2} & \textbf{88.5} \\
\Xhline{1.0pt}
\end{tabular}}
\caption{Comparison ($\%$) of the proposed PRGC method with the prior works. \textbf{Bold} marks the highest score, \underline{underline} marks the second best score and $\ddagger$ marks the results reported by the original papers.}
% and $\S$ marks results which are re-implementation by official code.
\label{main result}
\end{table*}

\section{Experiments}
\subsection{Datasets and Experimental Settings}
%To make a fair and comprehensive comparison with previous works, following \citet{yu2020joint} and \citet{wang2020tplinker}, we evaluate our model on two public datasets: NYT \cite{Riedel2010ModelingRA} and WebNLG \cite{gardent-etal-2017-creating} that both have two versions, respectively, and we denote the different versions as NYT*, NYT and WebNLG*, WebNLG.
%Each dataset has two versions and we denote the different versions as NYT*, NYT and WebNLG*, WebNLG.
%The first version NYT* and WebNLG* annotate the last word of entities, 
%and is used by \cite{zeng-etal-2018-extracting}\cite{zeng-etal-2019-learning}\cite{fu-etal-2019-graphrel}\cite{nayak2019effective}\cite{DBLP:journals/corr/abs-1909-03227}; 
%while the second version NYT and WebNLG annotate the whole entity span.

For fair and comprehensive comparison, we follow \citet{yu2020joint} and \citet{wang2020tplinker} to evaluate our model on two public datasets NYT \cite{Riedel2010ModelingRA} and WebNLG \cite{gardent-etal-2017-creating}, both of which have two versions, respectively. 
We denote the different versions as NYT*, NYT and WebNLG*, WebNLG. Note that NYT* and WebNLG* annotate the last word of entities, while NYT and WebNLG annotate the whole entity span.
%and is used by \cite{zheng-etal-2017-joint}\cite{Bekoulis_2018}\cite{yuan2020relation}. 
The statistics of the datasets are described in Table \ref{statistics}. Following \citet{DBLP:journals/corr/abs-1909-03227}, we further characterize the test set w.r.t. the overlapping patterns and the number of triples per sentence.

%Following \citet{DBLP:journals/corr/abs-1909-03227}, we split the test set by overlapping patterns and the number of triples for further study. The statistics are described in Table \ref{statistics}.

Following prior works mentioned above, an extracted relational triple is regarded as correct only if it is an exact match with ground truth, which means the last word of entities or the whole entity span (depending on the annotation protocol) of both subject and object and the relation are all correct. Meanwhile, we report the standard micro Precision (Prec.), Recall (Rec.) and F1-score for all the baselines. The implementation details are shown in Appendix \ref{imple details}.

We compare PRGC with eight strong baseline models and the state-of-the-art models CasRel~\cite{DBLP:journals/corr/abs-1909-03227} and TPLinker~\cite{wang2020tplinker}. All the experimental results of the baseline models are directly taken from \citet{wang2020tplinker} unless specified.

\begin{table}[ht]
\renewcommand\arraystretch{1.2}
\centering
\scalebox{0.83}{
\begin{tabular}{l|lcccc}
\Xhline{1.0pt}
\multicolumn{2}{c}{Model} & Normal & SEO & EPO & SOO\\
\Xhline{0.6pt}
\parbox[t]{2mm}{\multirow{5}{*}{\rotatebox[origin=c]{90}{NYT*}}} & OrderCopyRE & 71.2 & 69.4 & 72.8 & - \\
& ETL-Span & 88.5 & 87.6 & 60.3 & - \\
& CasRel & 87.3 & 91.4 & 92.0 & 77.0$\S$ \\
& TPLinker & 90.1 & 93.4 & 94.0 & \textbf{90.1}$\S$ \\
%\Xhline{0.6pt}
& PRGC & \textbf{91.0} & \textbf{94.0} & \textbf{94.5} & 81.8 \\
\Xhline{0.6pt}
\parbox[t]{2mm}{\multirow{5}{*}{\rotatebox[origin=c]{90}{WebNLG*}}} & OrderCopyRE & 65.4 & 60.1 & 67.4 & - \\
& ETL-Span & 87.3 & 91.5 & 80.5 & - \\
& CasRel & 89.4 & 92.2 & 94.7 & 90.4$\S$ \\
& TPLinker & 87.9 & 92.5 & 95.3 & 86.0$\S$ \\
%\Xhline{0.6pt}
& PRGC & \textbf{90.4} & \textbf{93.6} & \textbf{95.9} & \textbf{94.6} \\
\Xhline{1.0pt}
\end{tabular}}
\caption{F1-score ($\%$) of sentences with different overlapping patterns. \textbf{Bold} marks the highest score and $\S$ marks results obtained by official implementations.}
\label{patterns table}
\end{table}

\begin{table}[ht]
\renewcommand\arraystretch{1.2}
\centering
\scalebox{0.72}{
\begin{tabular}{l|lccccc}
\Xhline{1.0pt}
\multicolumn{2}{c}{Model} & $N = 1$ & $N = 2$ & $N = 3$ & $N = 4$ & $N \geq 5$ \\
\Xhline{0.6pt}
\parbox[t]{2mm}{\multirow{5}{*}{\rotatebox[origin=c]{90}{NYT*}}} & OrderCopyRE & 71.7 & 72.6 & 72.5 & 77.9 & 45.9 \\
& ETL-Span & 88.5 & 82.1 & 74.7 & 75.6 & 76.9 \\
& CasRel & 88.2 & 90.3 & 91.9 & 94.2 & 83.7 \\
& TPLinker & 90.0 & 92.8 & 93.1 & \textbf{96.1} & 90.0 \\
%\Xhline{0.6pt}
& PRGC & \textbf{91.1} & \textbf{93.0} & \textbf{93.5} & 95.5 & \textbf{93.0} \\
\Xhline{0.6pt}
\parbox[t]{2mm}{\multirow{5}{*}{\rotatebox[origin=c]{90}{WebNLG*}}} & OrderCopyRE & 63.4 & 62.2 & 64.4 & 57.2 & 55.7 \\
& ETL-Span & 82.1 & 86.5 & 91.4 & 89.5 & 91.1 \\
& CasRel & 89.3 & 90.8 & 94.2 & 92.4 & 90.9 \\
& TPLinker & 88.0 & 90.1 & 94.6 & 93.3 & 91.6 \\
%\Xhline{0.6pt}
& PRGC & \textbf{89.9} & \textbf{91.6} & \textbf{95.0} & \textbf{94.8} & \textbf{92.8} \\
\Xhline{1.0pt}
\end{tabular}}
\caption{F1-score ($\%$) of sentences with different numbers of triples where $N$ is the number of triples in a sentence. \textbf{Bold} marks the highest score.}
\label{number of triples}
\end{table}

\subsection{Experimental Results}
In this section, we present the overall results and the results of complex scenarios, while the results on different subtasks corresponding to different components in our model are described in Appendix \ref{r on dif subs}.
%We compare our model with the several models as baselines: \textbf{NovelTagging}\cite{zheng-etal-2017-joint}, %\textbf{CopyRE}\cite{zeng-etal-2018-extracting}, \textbf{MultiHead}\cite{Bekoulis_2018}, \textbf{GraphRel}\cite{fu-etal-2019-graphrel}, %\textbf{OrderCopyRE}\cite{zeng-etal-2019-learning}, \textbf{ETL-span}\cite{yu2020joint}, \textbf{WDec}\cite{nayak2019effective}, %\textbf{RSAN}\cite{yuan2020relation} and \textbf{CasRel}\cite{DBLP:journals/corr/abs-1909-03227}, \textbf{TPLinker}\cite{wang2020tplinker} %especially. Note that CasRel and TPLinker are strong state-of-the-art models, which has lower speed or more decoder parameters than our model.

\begin{table*}[ht]
\renewcommand\arraystretch{1.2}
\centering
\scalebox{0.8}{
\begin{tabular}{llccccc}
\Xhline{1.0pt}
Dataset & Model & Complexity & FLOPs (M) & Params$_{decoder}$ & Inference Time (1 / 24) & F1-Score \\
\Xhline{0.6pt}
\multirow{3}{*}{NYT*} & CasRel & \bm{$O(kn) \rightarrow O(n^2)$} & \textbf{15.05} & 75,362 & 24.2 / - & 89.6 \\
& TPLinker & $O(kn^2)$ & 1105.92 & 110,736 & 38.8 / 7.7 & 91.9  \\
& PRGC & $O(n^2)$ & 32.60 & \textbf{66,085} & \textbf{13.5 / 4.4} & \textbf{92.6} \\
\Xhline{0.6pt}
\multirow{3}{*}{WebNLG*} & CasRel & $O(kn^2)$ & 105.37 & 527,534 & 30.5 / - & 91.8 \\
& TPLinker & $O(n^3)$ & 7879.68 & 788,994 & 41.7 / 13.2 & 91.9  \\
& PRGC & \bm{$O(n^2)$} & \textbf{33.75} & \textbf{409,534} & \textbf{14.4 / 5.2} & \textbf{93.0} \\
\Xhline{1.0pt}
\end{tabular}}
\caption{Comparison of model efficiency on both NYT* and WebNLG* datasets. 
Results except F1-score (\%) of other methods are obtained by the official implementation with default configuration, and \textbf{bold} marks the best result. 
Complexity are the computation complexity, FLOPs and Params$_{decoder}$ are both calculated on the decoder, and we measure the inference time (ms) with the batch size of 1 and 24, respectively.}
\label{efficiency}
\end{table*}

\subsubsection{Overall Results}
Table \ref{main result} shows the results of our model against other baseline methods on four datasets. Our PRGC method outperforms them in respect of almost all evaluation metrics even if compared with the recent strongest baseline \cite{wang2020tplinker} which is quite complicated.

%It proves that our method has better performance on relational triple extraction task.
%It proves that our sequence tagging scheme with global correspondence instead of span based extraction method improves model result and robustness.

At the same time, we implement PRGC$_{Random}$ to validate the utility of our PRGC decoder, where all parameters of the encoder BERT are randomly initialized. 
The performance of PRGC$_{Random}$ demonstrates that our decoder framework (which obtains 7\% improvements than CasRel$_{Random}$) is still more competitive and robust than others even without taking advantage of the pre-trained BERT language model.

It is important to note that even though TPLinker$_{BERT}$ has more parameters than CasRel$_{BERT}$, it only obtains 0.1\% improvements on the WebNLG* dataset, and the authors attributed this to problems with the dataset itself.
%achieves a similar F1-score with CasRel$_{BERT}$ on the WebNLG* dataset, and the authors attributed this to problems with the dataset itself.
However, our model achieves a 10$\times$ improvements than TPLinker on the WebNLG* dataset and a significant promotion on the WebNLG dataset.
%more improvements on both versions of WebNLG. 
The reason behind this is that the relation judgement component of our model greatly reduces redundant relations particularly in the versions of WebNLG which contain hundreds of relations. 
In other words, the reduction in negative relations provides an additional boost compared to the models that perform entity extraction under every relation.

\subsubsection{Detailed Results on Complex Scenarios}

% \begin{table*}[ht]
% \renewcommand\arraystretch{1.2}
% \centering
% \scalebox{0.8}{
% \begin{tabular}{lccccccccc}
% \Xhline{1.0pt}
% \multirow{2}{*}{Model} & \multicolumn{4}{c}{NYT*} & \multicolumn{4}{c}{WebNLG*} \\
% \cmidrule(r){2-5}\cmidrule(r){6-9} & Normal & SEO & EPO & SOO & Normal & SEO & EPO & SOO\\
% \Xhline{0.6pt}
% OrderCopyRE & 71.2 & 69.4 & 72.8 & - & 65.4 & 60.1 & 67.4 & - \\
% ETL-Span & 88.5 & 87.6 & 60.3 & - & 87.3 & 91.5 & 80.5 & - \\
% CasRel & 87.3 & 91.4 & 92.0 & 77.0$\S$ & 89.4 & 92.2 & 94.7 & 90.4$\S$ \\
% TPLinker & 90.1 & 93.4 & 94.0 & \textbf{90.1}$\S$ & 87.9 & 92.5 & 95.3 & 86.0$\S$ \\
% \Xhline{0.6pt}
% FAST & \textbf{91.0} & \textbf{94.0} & \textbf{94.5} & 81.8 & \textbf{90.4} & \textbf{93.6} & \textbf{95.9} & \textbf{94.6} \\
% \Xhline{1.0pt}
% \end{tabular}}
% \caption{F1-score of sentences with different overlapping patterns. Bold marks the highest score, $\S$ marks results which are % re-implementation by official code, other results are all cited from \protect\cite{wang2020tplinker}.}
% \label{patterns table}
% \end{table*}

Following previous works~\citep{DBLP:journals/corr/abs-1909-03227, yuan2020relation, wang2020tplinker}, 
to verify the capability of our model in handling different overlapping patterns and sentences with different numbers of triples, we conduct further experiments on NYT* and WebNLG* datasets.

% \begin{table*}[ht]
% \renewcommand\arraystretch{1.2}
% \centering
% \scalebox{0.8}{
% \begin{tabular}{lccccccccccc}
% \Xhline{1.0pt}
% \multirow{2}{*}{Model} & \multicolumn{5}{c}{NYT*} & \multicolumn{5}{c}{WebNLG*} \\
% \cmidrule(r){2-6}\cmidrule(r){7-11} & N = 1 & N = 2 & N = 3 & N = 4 & N $\geq$ 5 & N = 1 & N = 2 & N = 3 & N = 4 & N $\geq$ 5 \\
% \Xhline{0.6pt}
% OrderCopyRE & 71.7 & 72.6 & 72.5 & 77.9 & 45.9 & 63.4 & 62.2 & 64.4 & 57.2 & 55.7 \\
% ETL-Span & 88.5 & 82.1 & 74.7 & 75.6 & 76.9 & 82.1 & 86.5 & 91.4 & 89.5 & 91.1 \\
% CasRel & 88.2 & 90.3 & 91.9 & 94.2 & 83.7 & 89.3 & 90.8 & 94.2 & 92.4 & 90.9 \\
% TPLinker & 90.0 & 92.8 & 93.1 & \textbf{96.1} & 90.0 & 88.0 & 90.1 & 94.6 & 93.3 & 91.6 \\
% \Xhline{0.6pt}
% FAST & \textbf{91.1} & \textbf{93.0} & \textbf{93.5} & 95.5 & \textbf{93.0} & \textbf{89.9} & \textbf{91.6} & \textbf{95.0} & \textbf{94.8} & % \textbf{92.8} \\
% \Xhline{1.0pt}
% \end{tabular}}
% \caption{F1-score of sentences with different number of triples. N is the number of triples in a sentence, bold marks the highest score, all % results are all cited from \protect\cite{wang2020tplinker}.}
% \label{number of triples}
% \end{table*}

As shown in Table \ref{patterns table}, our model exceeds all the baselines in all overlapping patterns in both datasets except the \textit{SOO} pattern in the NYT* dataset. Actually, the observation on the latter scenario is not reliable due to the very low percentage of \textit{SOO} in NYT* (i.e., 45 out of 8,110 as shown in Table \ref{statistics}).
As shown in Table \ref{number of triples}, the performance of our model is better than others almost in every subset regardless of the number of triples. In general, these two further experiments adequately show the advantages of our model in complex scenarios.

\section{Analysis}
\subsection{Model Efficiency}
\begin{figure}[ht]
\centering
\includegraphics[width=0.43\textwidth]{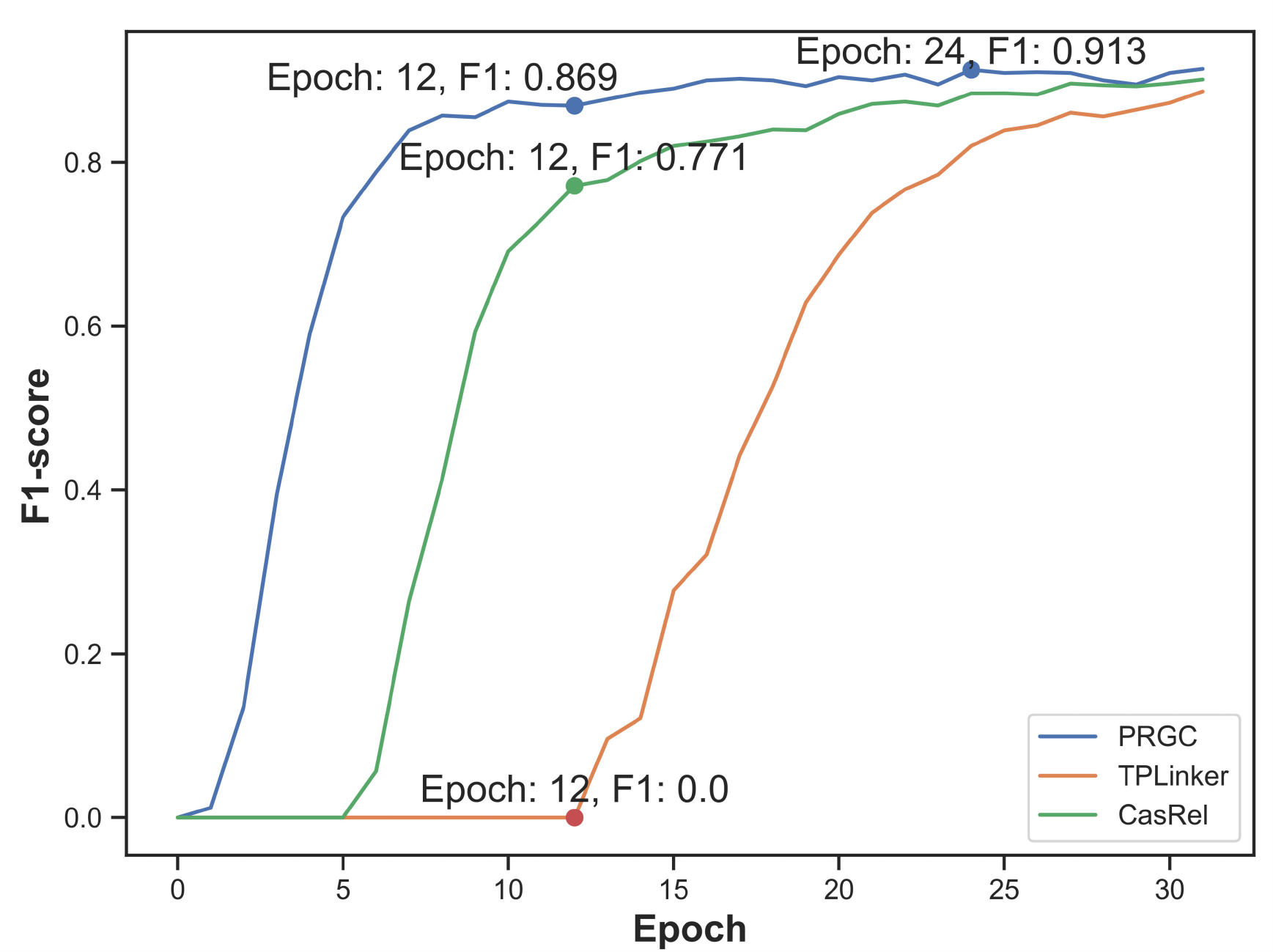}
\caption{F1-score with respect to the epoch number on the WebNLG* validation set of different methods. Results of CasRel and TPLinker are obtained by the official implementation with default configuration.}
\label{convergence}
\end{figure}

As shown in Table \ref{efficiency}, we evaluate the model efficiency with respect to \textit{Complexity}, floating point operations (\textit{FLOPs})~\cite{molchanov2017pruning}, parameters of the decoder (\textit{Params$_{decoder}$}) and \textit{Inference Time}\footnote{The \textit{FLOPs} and \textit{Params$_{decoder}$} are calculated via: https://github.com/sovrasov/flops-counter.pytorch.} of CasRel, TPLinker and PRGC in two datasets which have quite different characteristics in the size of relation set, the average number of relations per sentence and the average number of subjects per sentence.
All experiments are conducted with the same hardware configuration.
%which are different in $R$, $x_{rel}$ and $x_{sub}$ denote the size of relation set, the average number of relations per sentence and the average number of subjects per sentence respectively. \revise{The length of sentence $n$ is set to 100, $R$, $x_{rel}$ and $x_{sub}$ are 24, 2 and 2 in NYT* dataset and 171, 3 and 2 in WebNLG* dataset severally.} 
\begin{figure*}[ht]
\centering
\includegraphics[width=0.9\textwidth]{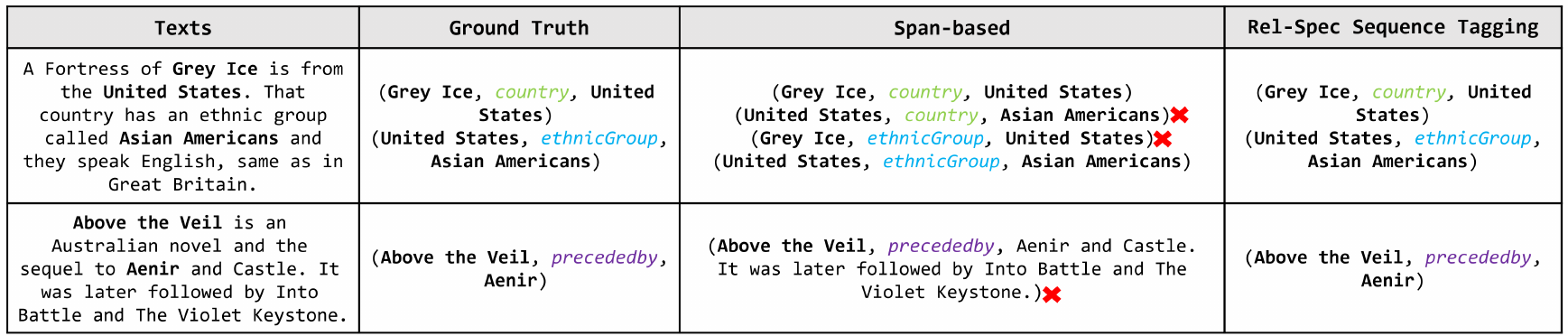}
\caption{Case study for the ablation study of Rel-Spec Sequence Tagging. Examples are from WebNLG*, and we supplement the whole entity span through WebNLG to facilitate viewing. The red cross marks bad cases, the correct entities are in \textbf{bold} and the correct relations are colored.}
\label{case study}
\end{figure*}
Because the number of subjects in a sentence varies, it is difficult for CasRel to predict objects in a heterogeneous batch, and it is restricted to set batch size to 1 in the official implementation \cite{wang2020tplinker}. For the sake of fair comparison, we set batch size to 1 and 24 to verify the single-thread decoding speed and parallel processing capability, respectively. 

The results indicate that the single-thread decoding speed of PRGC is 2$\times$ as CasRel and 3$\times$ as TPLinker, and our model is significantly better than TPLinker in terms of parallel processing.
Note that the model efficiency of CasRel and TPLinker decreases as the size of relation set increases but our model is not affected by the size of relation set, thus PRGC overwhelmingly outperforms both models in terms of all the indicators of efficiency in the WebNLG* dataset. 
Compared with the state-of-the-art model TPLinker, PRGC is an order of magnitude lower in \textit{Complexity} and the \textit{FLOPs} is even 200 times lower, thus PRGC has fewer parameters and obtains 3$\times$ speedup in the inference phase while the F1-score is improved by 1.1\%.
Even though CasRel has lower \textit{Complexity} and \textit{FLOPs} in the NYT* dataset, PRGC still has significant advantages and obtains a 5$\times$ speedup in the inference time and 3\% improvements in F1-score. 
Meanwhile, Figure \ref{convergence} proves our advantage in convergence rate. These all confirm the efficiency of our model.

% Figure \ref{convergence} shows the curves of F1-score on validation set varying with training epochs under different methods in the WebNLG* dataset which proves our advantages in convergence rate and model complexity.

\subsection{Ablation Study}

In this section, we conduct ablation experiments to demonstrate the effectiveness of each component in PRGC with results reported in Table \ref{ablation}. %Note that if we remove the \textit{Potential Relation Prediction} component from our model, the exposure bias will disappear, thus we can also verify the influence of exposure bias in PRGC.

\begin{table}[ht]
\renewcommand\arraystretch{1.2}
\centering
\scalebox{0.8}{
\begin{tabular}{l|lccc}
\Xhline{1.0pt}
\multicolumn{2}{c}{Model} & Prec. & Rec. & F1 \\
\Xhline{0.6pt}
\parbox[t]{2mm}{\multirow{4}{*}{\rotatebox[origin=c]{90}{NYT*}}} & \textbf{PRGC} & \textbf{93.3} & \textbf{91.9} & \textbf{92.6} \\
& \hspace{0.5em}--Potential Relation Prediction & 91.5 & 91.7 & 91.6  \\
& \hspace{0.5em}--Rel-Spec Sequence Tagging & 63.8 & 91.7 & 75.2 \\
& \hspace{0.5em}--Global Correspondence & 71.6 & 91.2 & 80.2 \\
\Xhline{0.6pt}
\parbox[t]{2mm}{\multirow{4}{*}{\rotatebox[origin=c]{90}{WebNLG*}}} & \textbf{PRGC} & \textbf{94.0} & \textbf{92.1} & \textbf{93.0} \\
& \hspace{0.5em}--Potential Relation Prediction & 80.0 & 88.2 & 83.9  \\
& \hspace{0.5em}--Rel-Spec Sequence Tagging & 33.2 & 91.3 & 48.7 \\
& \hspace{0.5em}--Global Correspondence & 55.9 & 91.6 & 69.4 \\
\Xhline{1.0pt}
\end{tabular}}
\caption{Ablation study of PRGC (\%).}
\label{ablation}
\end{table}

%\begin{table}[ht]
%\renewcommand\arraystretch{1.2}
%\centering
%\scalebox{0.63}{
%\begin{tabular}{lcccccc}
%\Xhline{1.0pt}
%\multirow{2}{*}{Model} & \multicolumn{3}{c}{NYT*} & \multicolumn{3}{c}{WebNLG*} \\
%\cmidrule(r){2-4}\cmidrule(r){5-7} & Prec. & Rec. & F1 & Prec. & Rec. & F1 \\
%\Xhline{0.6pt}
%\textbf{PRGC} & \textbf{93.3} & \textbf{91.9} & \textbf{92.6} & \textbf{94.0} & \textbf{92.1} & \textbf{93.0} \\
%\hspace{0.5em}--Potential Relation Prediction & 91.5 & 91.7 & 91.6 & 80.0 & 88.2 & 83.9  \\
%\hspace{0.5em}--Rel-Spec Sequence Tagging & 63.8 & 91.7 & 75.2 & 33.2 & 91.3 & 48.7\\
%\hspace{0.5em}--Global Correspondence & 71.6 & 91.2 & 80.2 & 55.9 & 91.6 & 69.4\\
%\Xhline{1.0pt}
%\end{tabular}}
%\caption{Ablation study of PRGC on both NYT* and WebNLG* datasets.}
%\label{ablation}
%\end{table}

\subsubsection{Effect of Potential Relation Prediction}
\label{ana on exposure}
We use each relation in the relation set to perform sequence tagging when we remove the \textit{Potential Relation Prediction} component to avoid the exposure bias.
As shown in Table \ref{ablation}, the precision significantly decreases without this component, because the number of predicted triples increases due to relations not presented in the sentences, especially in the WebNLG* dataset where the size of relation set is much bigger and brings tremendous relation redundancy. Meanwhile, with the increase of relation number in sentences, the training and inference time increases three to four times.
Through this experiment, the validity of this component that aims to predict a potential relation subset is proved, which is not only beneficial to model accuracy, but also to efficiency. 
%At the same time, this experiment proves that the \textit{Potential Relation Prediction} which introduces the exposure bias is beneficial overall.

%At the same time, this experiment proves that the exposure bias introduced in \em{Potential Relation Prediction} is beneficial overall.
%the advantages of \textit{Potential Relation Prediction} that introduces the exposure bias outweigh disadvantages.

%At the same time, this experiment proves that the advantages of \textit{Potential Relation Prediction} that introduces the exposure bias outweigh disadvantages. Therefore, we don't think it is necessary to pursue too much to avoid the exposure bias in end-to-end relational triple extraction model.

\subsubsection{Effect of Rel-Spec Sequence Tagging}
As a comparison for sequence tagging scheme, following \citet{DBLP:journals/corr/abs-1909-03227} and \citet{wang2020tplinker}, we perform binary classification to detect start and end positions of an entity with the span-based scheme. As shown in Table \ref{ablation}, span-based scheme brings significant decline of performance.

%Figure \ref{case study} show that span-based scheme trend to extract long entity and remember the position of entity rather than understand the semantics. 
%However, sequence tagging in PRGC performed perfectly in both cases due to the consideration of the interaction of each token(word or subword\footnote{WordPiece tokenization used by BERT\protect\cite{devlin2019bert}}) in an entity. Experiment result proves that the sequence tagging scheme is more robust and effective.

Through the case study shown in Figure \ref{case study}, we observe that the span-based scheme tends to extract long entities and identify the correct subject-object pairs but ignore their relation.
That is because the model is inclined to remember the position of an entity rather than understand the underlying semantics.
However, the sequence tagging scheme used by PRGC performs well in both cases, and experimental results prove that our tagging scheme is more robust and generalizable.
%due to the consideration of the interaction of each token(word or subword\footnote{WordPiece tokenization used by BERT\protect\cite{devlin2019bert}.}) in an entity. 

%there are rare such problems with the sequence tagging scheme in PRGC due to the consideration of the interaction of each token(word or subword\footnote{WordPiece tokenization used by BERT\protect\cite{devlin2019bert}}) in an entity. Experiment result proves that the sequence tagging scheme is a more robust, more generalized method for relational entity extraction.

\subsubsection{Effect of Global Correspondence}
For comparison, we exploit the heuristic nearest neighbor principle to combine the subject-object pairs which was used by \citet{zheng-etal-2017-joint} and \citet{yuan2020relation}. As shown in Table \ref{ablation}, the precision also significantly decreases without \textit{Global Correspondence}, because the number of predicted triples increases with many mismatched pairs when the model loses the constraint imposed by this component.
This experiment proves that the \textit{Global Correspondence} component is effective and greatly outperforms the heuristic nearest neighbor principle in the subject-object alignment task.

\section{Conclusion}

In this paper, we presented a brand-new perspective and introduced a novel joint relational extraction framework based on \textbf{P}otential \textbf{R}elation and \textbf{G}lobal \textbf{C}orrespondence, which greatly alleviates the problems of redundant relation judgement, poor generalization of span-based extraction and inefficient subject-object alignment. 
%Instead of extracting entities under all relations, we used the relation prediction component to predict potential relations;
%then a relation-specific sequence tagging scheme was adopted to eliminate the instability of span-based extraction; 
%for subject-object alignment, we designed the global correspondence component to accurately aligns subject-object pairs.
Experimental results showed that our model achieved the state-of-the-art performance in the public datasets and successfully handled many complex scenarios with higher efficiency.

% Each subtask solves several problems existing in previous work, 

\bibliographystyle{acl_natbib}
\bibliography{acl2021}

\clearpage
%\newpage
\appendix
\section*{Appendix}
\section{Overlapping Patterns}
\label{overlapping appendix}
\begin{figure}[ht]
\centering
\includegraphics[width=0.48\textwidth]{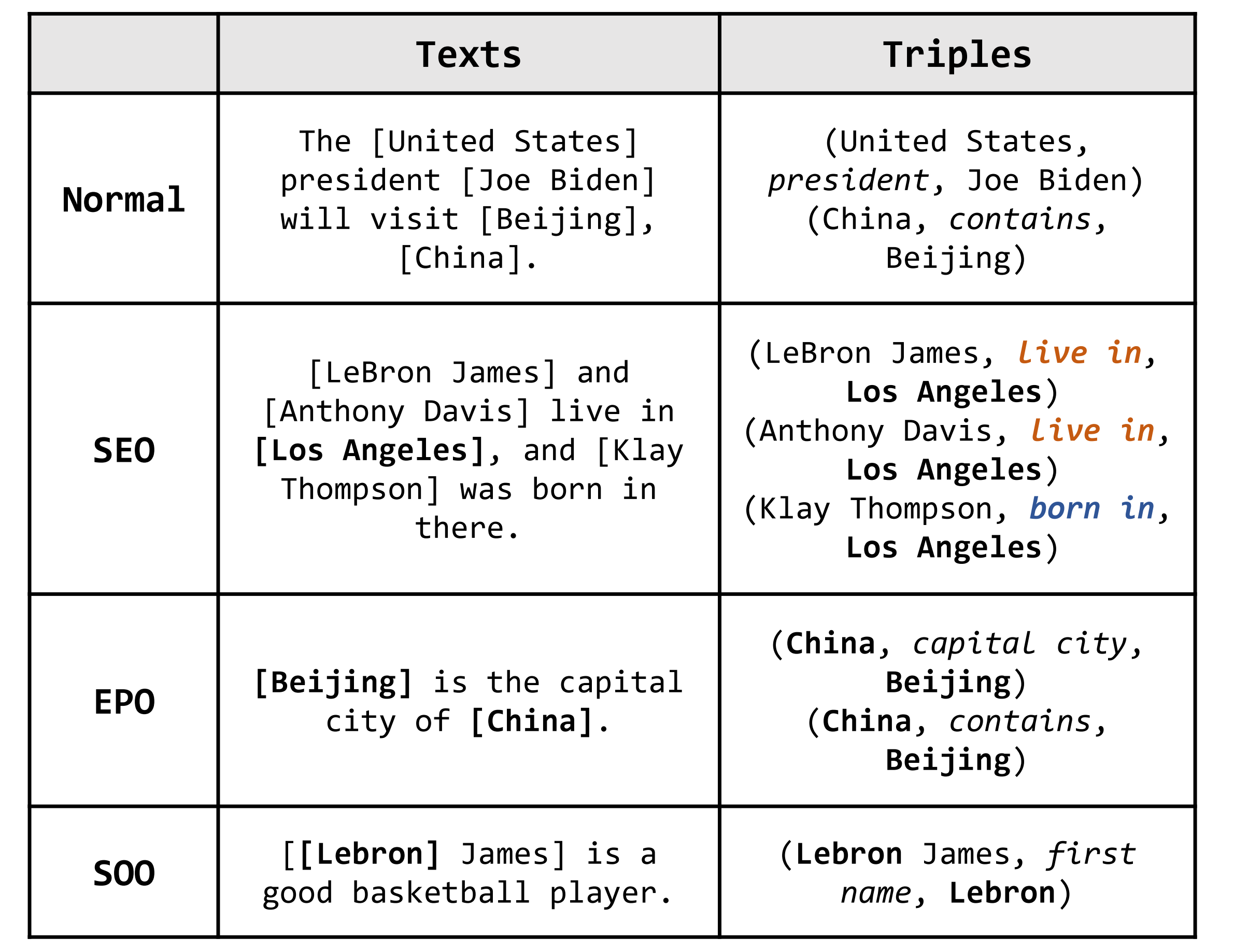}
\caption{Examples of the \textit{Normal}, \textit{Single Entity Overlap (SEO)}, \textit{Entity Pair Overlap (EPO)} and \textit{Subject Object Overlap (SOO)} patterns. The overlapping entities are in \textbf{bold}.}
\label{overlapping}
\end{figure}

As shown in Figure \ref{overlapping}, the \textit{Normal}, \textit{SEO} and \textit{EPO} patterns are usually mentioned in prior works~\citep{nayak2019effective, DBLP:journals/corr/abs-1909-03227, yuan2020relation, wang2020tplinker}, and \textit{SOO} is a special pattern we identified and addressed.

%\section{BERT Encoder}
%\label{bert}
%We use a pre-trained BERT model\cite{devlin2019bert} to encode the input sentence. Let $S = \{x_1, x_2, ..., x_n\}$ denote the input %sentence, $n$ is the number of tokens. The output of BERT is $Y_{BERT}(S) = \{v_1, v_2, ..., v_n | v_i \in \mathbb{R}^{d \times %1}\}$, $d$ is the embedding size. More detailed descriptions are shown in the original paper\cite{devlin2019bert}.

\section{Implementation Details}
\label{imple details}
We implement our model with PyTorch and optimize the parameters by Adam \cite{kingma2017adam} with batch size of 64/6 for NYT/WebNLG. The encoder learning rate for BERT is set as $5 \times 10^{-5}$, and the decoder learning rate is set as 0.001 in order to converge rapidly. We also conduct weight decay \cite{loshchilov2018fixing} with a rate of 0.01.

For fair comparison, we use the BERT-Base-Cased English model\footnote{Available at https://huggingface.co/bert-base-cased.} as our encoder, and set the max length of an input sentence to 100, which is the same as previous works~\citep{DBLP:journals/corr/abs-1909-03227, wang2020tplinker}. 
Our experiments are conducted on the workstation with an Intel Xeon E5 2.40 GHz CPU, 128 GB memory, an NVIDIA Tesla V100 GPU, and CentOS 7.2.
We train the model for 100 epochs and choose the last model. The performance will be better if the higher the threshold of \textit{Potential Relation Prediction} ($\lambda_1$), but tuning the threshold of \textit{Global Correspondence} ($\lambda_2$) will not help which is consistent with the analysis in Appendix \ref{r on dif subs}.

\section{Results on Different Subtasks}
\label{r on dif subs}
To further verify the results of the three subtasks in our new perspective and the performance of each component in our model, we present more detailed evaluations on NYT* and WebNLG* datasets in Table \ref{diff subtasks}.
%we implement three evaluation metrics by F1-score on NYT* and WebNLG* datasets, and the results are shown in Table \ref{diff subtasks}.

\begin{table}[ht]
\renewcommand\arraystretch{1.2}
\centering
\scalebox{0.675}{
\begin{tabular}{lcccccc}
\Xhline{1.0pt}
\multirow{2}{*}{Subtask} & \multicolumn{3}{c}{NYT*} & \multicolumn{3}{c}{WebNLG*} \\
\cmidrule(r){2-4}\cmidrule(r){5-7} & Prec. & Rec. & F1 & Prec. & Rec. & F1 \\
\Xhline{0.6pt}
Relation Judgement & 95.3 & \textbf{96.3} & 95.8 & 92.8 & \textbf{96.2} & 94.5 \\
Entity Extraction (Subject) & 81.2 & \textbf{95.5} & 87.8 & 69.4 & \textbf{96.3} & 80.7 \\
Entity Extraction (Object) & 82.8 & \textbf{95.8} & 88.8 & 72.1 & \textbf{95.7} & 82.2 \\
Subject-object Alignment & 94.0 & 92.3 & \textbf{93.1} & 96.0 & 93.4 & \textbf{94.7} \\
Combination of Above All & 93.3 & 91.9 & 92.6 & 94.0 & 92.1 & 93.0 \\
\Xhline{1.0pt}
\end{tabular}}
\caption{Evaluation (\%) of different subtasks on the NYT* and WebNLG* datasets. Each subtask corresponds to a component in our model. \textbf{Bold} marks the most important metric of each subtask.}
\label{diff subtasks}
\end{table}

\paragraph{Relation Judgement}
We evaluate outputs of the \textit{Potential Relation Prediction} component which are potential relations contained in a sentence. Recall is more important for this task because if a true relation is missed, it will not be recovered in the following steps.
We get high recall in this task and the results show that effectiveness of \textit{Potential Relation Prediction} component is not affected by the size of relation set.

\paragraph{Entity Extraction}
This task is related to the \textit{Relation-Specific Sequence Tagging} component, and we evaluate it as a Named Entity Recognition (NER) task with two types of entities: subjects and objects.
The predicted entities are from all potential relations of a sentence, and recall is more important for this task because most false negatives can be filtered out by \textit{Subject-object Alignment}. Experimental results show that we extract almost all correct entities, and it further proves that the influence of the exposure bias is negligible.

\paragraph{Subject-object Alignment}
This task is related to the \textit{Global Correspondence} component, and we just evaluate the entity pair in a triple and ignore the relation. 
Both recall and precision are important for this component, experimental results indicate that our alignment scheme is useful but still can be further improved, especially in the recall.
%The precision is more important because it is a filter for other components, which means the content of triple is provided by other two subtasks.
%Experimental results indicate that our alignment scheme is useful to other subtasks.

Overall, the combination of three components in our model accomplishes the relational triple extraction task with a fine-grained perspective, and achieves better and solid results. 
%As described in Section \ref{loss func}, model performance might be better after carefully tuning the weight of each loss of subtasks.

\end{document}